\documentclass[letterpaper]{article} 
\usepackage{aaai25}  
\usepackage{times}  
\usepackage{helvet}  
\usepackage{courier}  
\usepackage[hyphens]{url}  
\usepackage{graphicx} 
\usepackage{natbib}  
\usepackage{caption} 
\usepackage{tabularx} 
\usepackage{array}    
\usepackage{algorithm}
\usepackage{algorithmic}
\usepackage{amsmath}
\usepackage{xcolor}
\usepackage{newfloat}
\usepackage{listings}
\DeclareCaptionStyle{ruled}{labelfont=normalfont,labelsep=colon,strut=off} 
\floatstyle{ruled}
\newfloat{listing}{tb}{lst}{}
\floatname{listing}{Listing}
\nocopyright

\title{Memory-Augmented Agent Training for Business Document Understanding}
\author{
    Jiale Liu\textsuperscript{\rm 1},
    Yifan Zeng\textsuperscript{\rm 2},
    Malte Højmark-Bertelsen\textsuperscript{\rm 3},
    Marie Normann Gadeberg\textsuperscript{\rm 3},\\
    Huazheng Wang\textsuperscript{\rm 2}, 
    Qingyun Wu\textsuperscript{\rm 1}
}
\affiliations{
    \textsuperscript{\rm 1}Pennsylvania State University
    \textsuperscript{\rm 2}Oregon State University
    \textsuperscript{\rm 3}Beyond Work

}

\begin{document}

\maketitle

\begin{abstract}
Traditional enterprises face significant challenges in processing business documents, where tasks like extracting transport references from invoices remain largely manual despite their crucial role in logistics operations. While Large Language Models offer potential automation, their direct application to specialized business domains often yields unsatisfactory results. We introduce Matrix (Memory-Augmented agent Training through Reasoning and Iterative eXploration), a novel paradigm that enables LLM agents to progressively build domain expertise through experience-driven memory refinement and iterative learning. To validate this approach, we collaborate with one of the world's largest logistics companies to create a dataset of Universal Business Language format invoice documents, focusing on the task of transport reference extraction\footnote{The dataset contains sensitive information from corporate customers and therefore cannot be released. We release an anonymized subset to facilitate research in this field. Data available at \color{magenta}https://tinyurl.com/3pakk9t4}. Experiments demonstrate that Matrix outperforms prompting a single LLM by 30.3\%, vanilla LLM agent by 35.2\%. We further analyze the metrics of the optimized systems and observe that the agent system requires less API calls, fewer costs and can analyze longer documents on average. Our methods establish a new approach to transform general-purpose LLMs into specialized business tools through systematic memory enhancement in document processing tasks.

\end{abstract}

%

\section{Introduction}
Combing through large quantities of unstructured data remains a widespread challenge in enterprise operations, particularly in finance functions where efficient invoice processing represents a growing competitive advantage. Despite the prevalent adoption of digital invoicing, many organizations still grapple with the labor-intensive and error-prone task of manually extracting crucial identifiers from business transactions. For logistics companies manual extraction not only slows down operations but also introduces the potential for human error, leading to misrouted shipments and customer dissatisfaction.

Large Language Models have demonstrated remarkable capabilities in natural language understanding and processing~\citep{achiam2023gpt,yang2024qwen2,dubey2024llama}. Building upon these capabilities, agent-based systems that leverage LLMs for multi-step reasoning have emerged as a promising paradigm~\citep{hu2024automated,wu2024mathchat,wu2023autogen}. However, these systems face significant challenges when adapting to specialized contexts without manual tuning, as LLMs are not specifically trained to reason over business documents. 

To address these limitations, we propose Matrix (Memory-Augmented agent Training through Reasoning and Iterative eXploration), a novel framework that enables LLM-based agents to learn and adapt from their experiences. Matrix incorporates a unique iterative self-refinement mechanism that allows agents to systematically improve their understanding of document structures and extraction patterns. The agents go through task exploration and optimization to iteratively improve their insights on the general task structure. The distilled actionable insights will be saved into a long-term memory of the agents and will guide future task solving attempts, leading to agent systems with better strategic reasoning and task solving capability. 

In collaboration with Kuehne+Nagel, one of the world's leading logistics companies, we collect data from real-world business invoice documents and benchmark Matrix on the dataset. Specifically, we focus on the challenge of extracting transport reference numbers, which directly impact shipment routing and supply chain visibility. To facilitate broader research in this domain, we also introduce the first open-source dataset of anonymized business invoices. This benchmark poses a unique opportunity to evaluate agent system's adaptive learning capabilities in critical business context.


Our experimental results demonstrate Matrix's effectiveness after several epochs of optimization. The system achieves significant improvements through iterative learning and demonstrates strong performance. When compared to previous baselines, Matrix achieves 30.3\% performance improvement compared to chain-of-thought prompting~\citep{wei2022chain}, surpasses vanilla LLM agents~\citep{wu2023autogen} by 35.2\%, reflexion~\citep{shinn2024reflexion} by 27.28\%. Further analysis on the system reveals that the learning process leads to significantly lower latency and costs, with a stronger capability in processing longer documents.

The key contributions of this work are threefold:
\begin{enumerate}
\item We propose a novel agent-based system for document reasoning and information extraction. It serves as a strong baseline in business cases.
\item We introduce Matrix, a novel paradigm that enables LLM-based agents to systematically learn and adapt to novel tasks through iterative self-refinement and experience-based memory updating.
\item We present the first open-source benchmark for evaluating document understanding systems on standardized business documents, facilitating reproducible research and practical applications in enterprise settings. This benchmark provides a challenging and practically relevant test for document reasoning and information retrieval capabilities in business contexts using agentic networks. 
\end{enumerate}

\section{Related Work}
\label{others}
\textbf{Business Document Reasoning Benchmarks.} 
While general-domain question answering datasets have driven advances in natural language understanding~\citep{abujabal-etal-2019-comqa,rajpurkar-etal-2016-squad,yang2018hotpotqa,joshi-etal-2017-triviaqa,talmor18compwebq,bajaj2016ms,kwiatkowski2019natural}, they fail to capture the structured, transactional nature of business documents. Existing research on business document understanding mostly focus on vision-based information extraction from scanned copies~\citep{harley2015evaluation,riba2019table,zhong2019publaynet,antonacopoulos2009realistic}, which emphasize layout understanding over semantic reasoning. These datasets, while valuable for OCR and structure recognition tasks, fail to capture the domain-specific patterns essential for business document processing. Moreover, text-based researches tend to use proprietary, undisclosed datasets~\citep{hamdi2021information,krieger2021information,palm2017cloudscan,tarawneh2019invoice}, limiting reproducibility and real-world applicability. To address these limitations, we present a publicly available business invoice dataset derived from real-world transactions to facilitate research on practical business scenarios.

\textbf{Prompt Optimization.} Prompt optimization is a popular paradigm for maximizing LLM's performance to novel tasks without expensive model tuning by finding a optimal task prompt~\citep{zhou2022large,pryzant2023automatic,cheng2023black,prasad2022grips}. In-context learning emerges as a prominent paradigm, where a set of input-output pairs is provided as few shot examples to the LLM~\citep{min2021metaicl,dong2022survey,brown2020language}. By automatically retrieving demonstrations from training set~\citep{zhao21calibrate,lu2021fantastically,liu2021makes} or from adaptively annotated samples~\citep{zhang2023ideal,wu2022self,su2022selective}, they primarily treat demonstrations as static examples rather than distilled insights. In contrast to this line of work, our proposed paradigm aims to distill trajectories into generalizable heuristics.

\textbf{Agent Learning.} There has been efforts in exploring inference time performance boost since the emergence of Large Language Models~\citep{shinn2024reflexion, madaan2024self,yao2023react,yao2024tree,sumers2023cognitive,wei2022chain,zhou2022least,guo2024embodied}. Recent works have extended this paradigm to agentic systems. 
Some works represent and learn the optimal workflow of agentic systems in the form of complex graphs~\citep{zhuge2024language,wu2024stateflow}, code~\citep{hu2024automated}, memory~\citep{wang2024agent} and trees~\citep{zhang2024aflow} to improve the system's performance on complex tasks, while others learns reusable tools~\citep{zhangoffline,cai2023large,qian2023creator,yuan2023craft} and experience~\citep{zhao2024expel,wang2024agent} for agentic systems. Different from previous memory optimization based works, our proposed approach does not rely on any pre-defined in-context examples~\citep{zhao2024expel} and meanwhile guarantee the diversity and robustness of the learned memory.

\section{Matrix: Memory-Augmented agent Training through Reasoning and Iterative eXploration}

\begin{figure*}[t]
    \centering
    \includegraphics[width=0.6\linewidth]{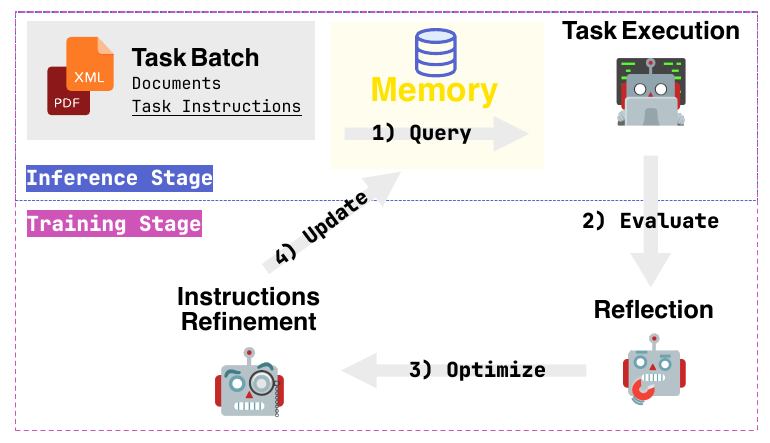}
    \caption{The training and inference pipeline of Matrix.}
    \label{fig:matrix-main-figure}
\end{figure*}

In this section, we begin with presenting the formulation of document reasoning, then introduce the framework of Matrix.
\subsection{Problem Statement}
\label{Method}
We begin with the formal definition of the document reasoning problem. In this problem, we assume the existence of a dataset for a given  task, which consists of  \(N\) instances of document, query, and answer triplets:
\begin{equation}
    \mathcal{D} = \{(d_i, q_i, a_i)\}_{i=1}^N
\end{equation}
where \( d_i \) represents a document, \( q_i \) is an associated query, and \( a_i \) is the corresponding correct answer. In the context of transport reference extraction from business documents, this formulation naturally maps to our task of interest: each document \(d_i\) represents a UBL invoice, the query \(q_i\) requests the location of the transport reference number, and the answer \(a_i\) is the correct reference number. This mapping allows us to frame the challenge of invoice processing as a structured document reasoning problem while maintaining the generality of our approach.

The overall goal in document reasoning is to train a system that learns a function \( f: D \times Q \rightarrow A \), where \( D \), \( Q \), \( A \)  represents the space of documents, queries, and answers respectively. For a given document \( d_i \) and query \( q_i \), the agent generates an answer \( \hat{a_i} \). The system's performance is evaluated based on the alignment between \( \hat{a_i} \) and \( a_i \).

We consider an agent with a long-term memory module \( M \), which stores useful contextual information for reasoning. For a given query \( (d, q) \), the agent solves the task in an iterative manner. At each timestep \( t \), the agent observes the current state \( o_t \), takes an action \( a_t \) (which may involve interacting with the document or recalling information from memory), and then receives an updated observation \( o_{t+1} \). This sequence of interactions produces a trajectory of observations and actions: \( \tau = (o_0, a_0, o_1, a_1, \dots, o_T, a_T) \). Finally, the agent produces the final predicted answer \( \hat{a}_i \) through an answer extraction module \( g(\tau) \), which takes into account the full trajectory \( \tau \) of observations and actions across timesteps. Thus, the system can be viewed as generating the prediction:
\begin{equation}
    \hat{a}_i = g(\tau) = f(d_i, q_i | M)
\end{equation}

The formal goal of optimizing the agent's reasoning ability can be described via the objective:
\begin{equation}
\label{objective}
    \arg\min_M \mathbf{E}_{(d_i, q_i, a_i)\sim\mathcal{D}}[\mathcal{L}(f(d_i, q_i|M), a_i)]
\end{equation}
where \( \mathcal{L} \) measures the discrepancy between generated answer and the ground truth. In our case, \( \mathcal{L} \) is defined using an exact match comparison.

\subsection{Proposed Method}
To optimize the objective in Eq.\ref{objective}, we follow the paradigm of supervised learning and develop the dataset into train set \( \mathcal{D}_{train} \) and test set \( \mathcal{D}_{test} \) to approximate the inaccessible real data distribution. We train an optimal memory and test it on test task set. The pipeline is illustrated in Figure \ref{fig:matrix-main-figure}.

\textbf{Trajectory Sampling.} We perform optimization by progressively updating the memory module \( M \) over multiple epochs. During each epoch, we sample a mini-batch of tasks from \( \mathcal{D}_{train} \). For each sampled task \( (d_i, q_i, a_i) \), the agent interacts through a sequence of actions over time, forming a trajectory \( \tau_i = (o^i_0, a^i_0, o^i_1, a^i_1, \dots, o^i_T, a^i_T) \), where \( o^i_t \) and \( a^i_t \) denote the observation and the action at timestep \( t \) for the \( i \)-th task.

The trajectory continues until the agent either reaches a solution or the interaction exceeds a predefined maximum number of steps. We enforce a upper limit \(T_{max}\) to the total number of steps per task. This can constrain the token count in each trajectory, and allow us to increase the mini-batch size during optimization. By doing so, the optimizer can process more diverse tasks within a single batch, which increases the model's exposure to varied problem domains, leading to a more robust and generalizable memory \( M \).

\textbf{Reflection.} Despite the impressive reasoning capabilities of LLM-based agents, they are prone to issues such as hallucination~\citep{ji2023towards}, factual errors~\citep{wang2023survey}, and reasoning failures~\citep{huang2023large}. In a limited number of steps, it is hard for agent to autonomously recognize and rectify its own mistakes without explicit self-correction mechanisms or auxiliary prompts~\citep{huang2023large,jiang2024self,pan2023automatically,kamoi2024can,song2024adaptive}. To overcome this limitation, we introduce a Reflector module that operates as a post hoc evaluator of the trajectory. The reflector is provided with the trajectory \( \tau_i\) and the ground truth answer \(y_i\). It's task is to label the solution as "Correct/Incorrect" and identify the key steps where reasoning errors occurred or correct decisions were made that led to the proper solution. The reflection process is described as:

\begin{equation}
    r_i = \text{LM}_{\text{reflect}}(\tau_i, y_i)
\end{equation}
where \(\text{LM}_{\text{reflect}}(\cdot)\) represents prompting a LLM to evaluate the trajectory. The reflection phase provides insights for refining the memory and improving the agent's task-solving strategy.

\textbf{Optimization.} To obtain an optimal solution in Eq.\ref{objective}, an optimizer is needed that can generate new solutions based on performance measurement. The optimization problem operates on the space of natural languages, which perfectly paves the way for borrowing Large Language Models' exceptional capability in natural language understanding~\citep{yang2023large,zhangoffline}. We propose a meta-optimizer with Large Language Models as the backend. The optimizer takes in three parts of information: (1) The execution trajectory of the agents in solving the tasks in the current training mini-batch (2) The assessment of each trajectory provided by the evaluator. (3) The current memory the agents operate on. Although the memory can be accessed by reading the prompt in the trajectory, we explicitly provide the current memory to let the optimizer progressively update the instruction based on the assessments for improvement. The optimization process can be formulated as:

\begin{equation}
    M_{i+1} = \text{LM}_{\text{optim}}(\tau, y, M_i)
\end{equation}
where $\text{LM}_{\text{optim}}(\cdot)$ denotes the process of LLM-based optimization. 


\section{Evaluations}
\subsection{Experiment Setup}

\textbf{Agent system.} We experiment with refining a two-agent system. In the system, an assistant powered by LLM receives and analyzes the task and suggests code for execution, and a user proxy automatically executes code and provide the result. The conversation will end once a final answer is reached by assistant or the max number of conversation turns is reached. The system is implemented by AG2\footnote{https://github.com/ag2ai/ag2} (formerly AutoGen)~\citep{wu2023autogen}.

\textbf{Experiment Dataset.}  We evaluate our system on a collection of real-world Universal Business Language invoice documents, developed in cooperation with Kuehne+Nagel, one of the world's largest logistics companies. The primary task is to extract transport reference number from these documents. The dataset contains 764 valid invoice document and transport reference pairs. While the real world dataset contains sensitive customer data and cannot be released publicly, we provide an anonymized subset of the dataset and include the evaluation results in appendix.

The dataset presents several challenging characteristics that make it an ideal testbed for evaluating iterative learning capabilities. First, it requires specialized domain knowledge of business documents and terminology not commonly found in general language model training. Second, the hierarchical structure of UBL documents and the significant variability in format and identification patterns pose substantial extraction challenges. Additionally, as a novel benchmark without prior literature coverage, this dataset offers unique opportunities to assess agents' adaptive learning abilities in a practical, high-stakes business context.

\textbf{Evaluation Protocol.} We report the success rate of the agent. The agent system is required to output the final result into a standardized format. The task is considered success when the output transport reference exactly matches the ground truth label. Given the diverse range of possible edge cases in the outputs, manually creating an extraction and comparison module may not cover all scenarios. Therefore, we employ a LLM judge~\citep{zheng2023judging} to compare the agent's output with the ground truth.

\textbf{Baselines for Comparison.} We compare performance with previous works, including prompting LLM using Chain of Thought~\citep{wei2022chain}, Reflection~\citep{shinn2024reflexion}, and agent system with no memory~\citep{wu2023autogen}. For a detailed description of the implementation details of the baselines, please refer to appendix \ref{appendix:implementation}.

\subsection{Main Results}
\label{main}


\begin{figure}[htbp]
    \centering
    \includegraphics[width=0.9\linewidth]{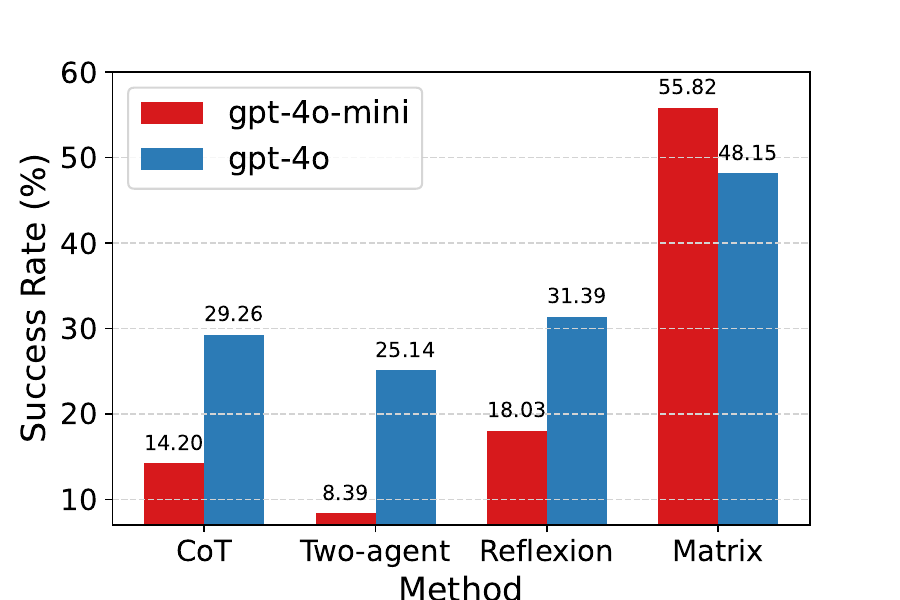}
    \caption{Comparison between Matrix and baselines with \texttt{gpt-4o} and \texttt{gpt-4o-mini} as backbone model. Matrix leverages \texttt{gpt-4o} for optimization in both cases. Surprisingly, \texttt{gpt-4o-mini} performs better after optimization.}
    \label{fig:main_result}
\end{figure}

We randomly select 60 samples from the dataset for training, the remaining 704 samples are reserved for testing performance. The maximum number of conversational turns between the assistant agent and the user proxy was capped at 5. The batch size was set to 14. That is, for each epoch, 14 tasks are sampled from training set and used for optimization. Since individual trajectories can become excessively lengthy, making the input to optimizer LLM larger than its context limit, we addressed this by truncating the batch to include only the largest number of trajectories that fit within the context limit. We set the backbone of the optimizer to \texttt{gpt-4o}. To maintain a fair comparison, we use \texttt{gpt-4o} for providing verbal feedback while running reflexion.

As shown in Figure \ref{fig:main_result}, Matrix performs the best across all previous benchmarks. Without the memory module, vanilla agent cannot perform well without any prior knowledge, even performing worse than directly prompting the language model with chain-of-thought. In this case, equipping the agent with code interpreter actually hampers the performance, as the agent will over rely on writing code to extract strings rather than using natural language to reason. With the proposed Matrix mechanism, the agent's task solving attempt is correctly guided by the long-term memory. The performance nearly doubles and outperforms all other methods.

\begin{figure}[htbp]
    \centering
    \includegraphics[width=0.8\linewidth]{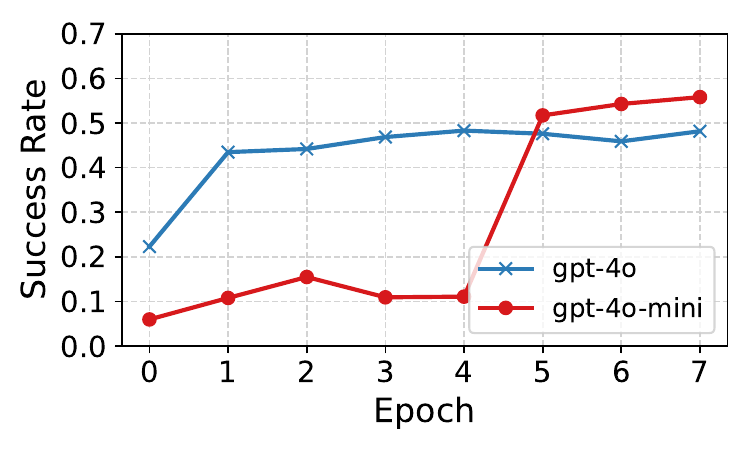}
    \caption{Success rate comparison between agent with \texttt{gpt-4o} and \texttt{gpt-4o-mini} as backbone over epochs.}
    \label{fig:epoch}
\end{figure}

Figure \ref{fig:epoch} demonstrates the performance of the agent with respect to the number of optimization epochs. The performance shows a steady rise across all epochs. With no memory, the agent system's performance relies entirely on the capability of backbone LLM.

\subsection{Analysis of the Optimized System}

In this section, we analyze the metrics of the system across the optimization of memory across epochs.
Specifically, we explore three metrics: (1) average api calls it takes to correctly solve a question. This affects the overall latency of the agent system. (2) average cost it takes to correctly solve a question. (3) the distribution of the length of the successfully solved documents.

\begin{figure}[htbp]
    \centering
    \includegraphics[width=0.7\linewidth]{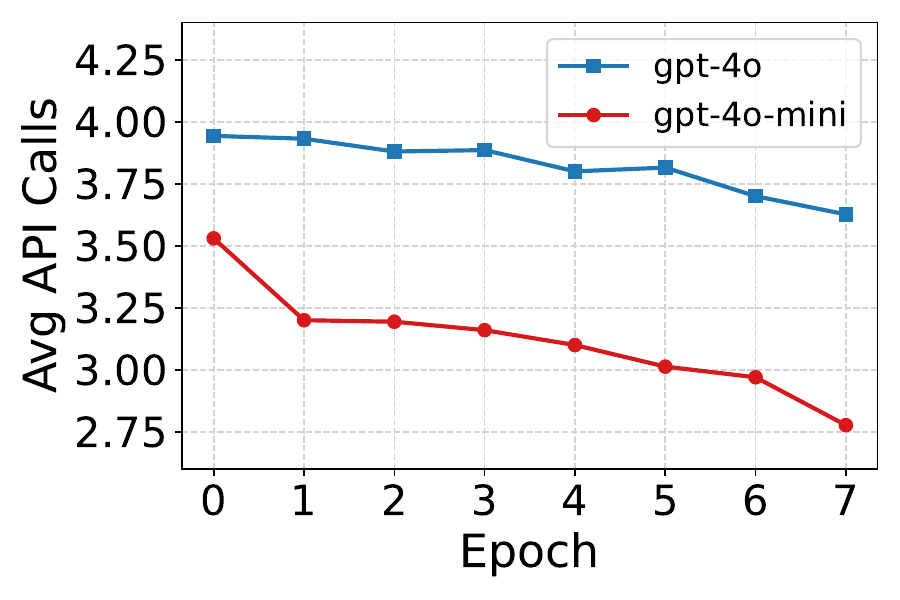}
    \caption{Comparison of average number of API calls it takes to solve a task. The average number decreases steadily as the training goes on.}
    \label{fig:comp_api}
\end{figure}

\textbf{Analysis of the average number of API calls.} As shown in Figure \ref{fig:comp_api}, the agent system exhibits a notable decrease in the average number of API calls required to solve the document question-answering task. After equipping with the optimized memory, the average number of API calls reduces about 8.12\% for \texttt{gpt-4o} backed agent, 21.3\% for \texttt{gpt-4o-mini} backed agent. This reduction indicates that the system is able to handle the task more efficiently with fewer calls, reducing sources of latency but reaching a better performance.

\begin{figure}[htbp]
    \centering
    \includegraphics[width=0.8\linewidth]{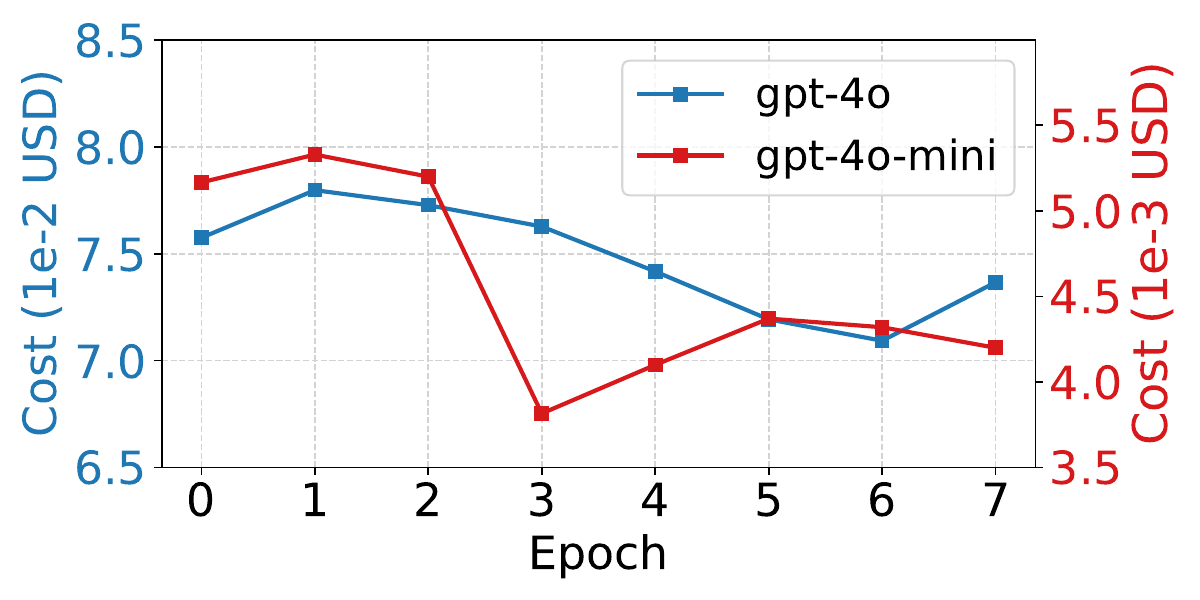}
    \caption{Average cost of API calls after each epoch. The cost shows a decreasing trend as training goes on.}
    \label{fig:comp_cost}
\end{figure}

\textbf{Analysis of the cost.} We plot the average cost of successfully solving a task after each epoch in Figure \ref{fig:comp_cost}. The cost increases during the early stages of optimization. This rise is attributed to the initial memory initialization and optimization phase. During this phase, while the memory is introduced to guide problem-solving, its content is not yet fully refined. As a result, the agent writes more lengthy code for trial and subsequently leading to higher API costs. As optimization goes on, the memory gets gradually refined and more comprehensive, the agent can solve the task more efficiently with less tokens. The slight uptick in at the end of the optimization suggests an adaptive process, where some rebalancing occurs after significant cost reduction. From a broader perspective, the overall trajectory suggests that optimization effectively lowers the long-term computational cost in solving tasks.

\begin{figure}[htbp]
    \centering
    \includegraphics[width=0.8\linewidth]{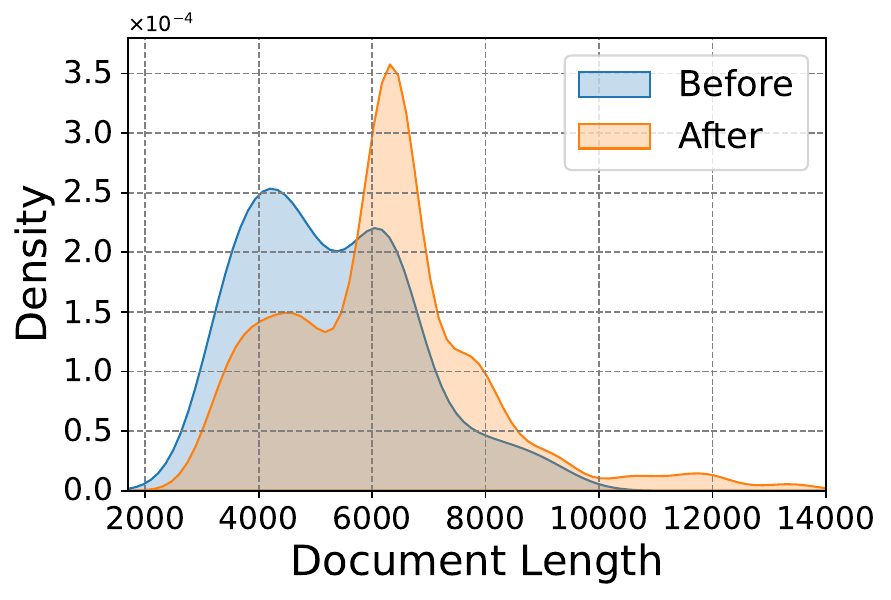}
    \caption{Distribution of successfully analyzed document lengths before and after optimization with \texttt{gpt-4o} as backbone.}
    \label{fig:comp_length}
\end{figure}

\textbf{Analysis of the length of successfully solved documents.} As shown in Figure \ref{fig:comp_length}, the distribution of the successfully analyzed document lengths by \texttt{gpt-4o} backed agent is illustrated for two different stages of the agent's performance: before optimization and after optimization. The document length is measured by the total number of tokens after tokenization. The distribution of successfully analyzed document lengths shifts notably after optimization. Before training, the performance peaks for document lengths around 4000 and 6000 tokens, and significantly declines after 6000 tokens. After training, we observe a significant improvement, especially for longer documents, with the distribution peak shifting towards 6,000 tokens and an extended tail that stretches beyond. This pattern showcases the enhanced capacity of the agent system to handle larger documents after optimization, reflecting its improved robustness and processing efficiency for more complex and lengthy inputs. We observe similar pattern for \texttt{gpt-4o-mini} backed agent and therefore omit the plotting for brevity.

\subsection{Optimization with Weaker Language Model}
\label{weak}
In the main experiment, we leveraged \texttt{gpt-4o} as backbone for optimization. We next explore the performance with smaller language model used as optimizer. Following the protocol in Section \ref{main}, we set the backbone of optimizer and agents both to \texttt{gpt-4o-mini}. 

\begin{figure}[htbp]
    \centering
    \includegraphics[width=0.7\linewidth]{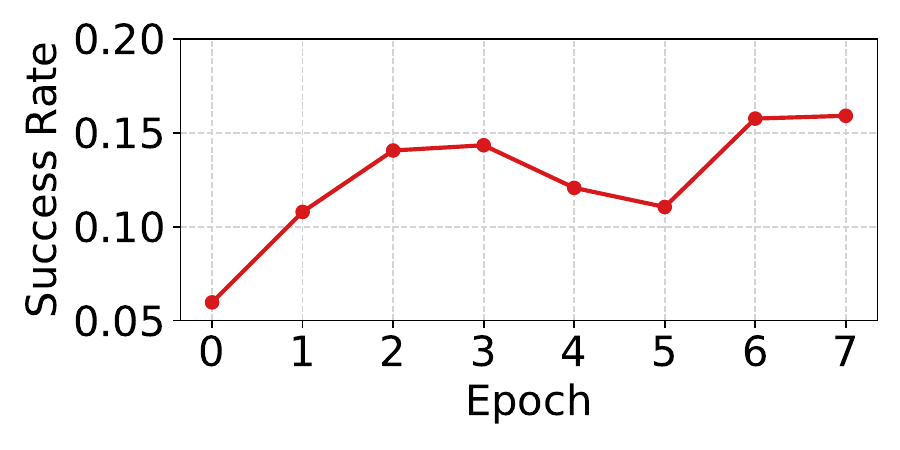}
    \caption{Performance of the optimized agent across iterations using \texttt{gpt-4o-mini} as optimizer. The performance rises in the beginning, showing steady improvement in success rate over the first three epochs, but soon begins to stabilize and reaches a slightly higher plateau. }
    \label{fig:comp-mini}
\end{figure}

Figure \ref{fig:comp-mini} demonstrates how the performance changes with respect to the number of training epochs. Although \texttt{gpt-4o-mini} has weaker language understanding capability, it can still understand the task and summarize reuable patterns from the task trajectories. However, it misses some regular expression patterns, leading to a less comprehensive optimized memory. Therefore, the final performance of the optimized system remains limited.

We then compare the performance between Matrix and Reflexion with \texttt{gpt-4o-mini} for optimization and providing verbal feedback, since they are both learning-based methods. The results, presented in Table \ref{tab:reflexion_mini}, indicate that Matrix consistently outperforms Reflexion. Notably, the online algorithmic nature of Reflexion necessitates collecting verbal feedback for agents through trial and error on a task-by-task basis, which is inefficient as the number of test tasks increases. In contrast, Matrix leverages a generalized pattern learned from trajectories, enabling it to enhance performance in a more cost-efficient manner.

\renewcommand{\arraystretch}{1.3}
\begin{table}[]
\centering
\begin{tabularx}{0.8\linewidth}{lXX}
\hline
Opt. LM & Reflexion & Matrix \\ \hline
\texttt{gpt-4o-mini}       & 12.64     & \textbf{15.90}  \\ \hline
\texttt{gpt-4o}            & 18.03     & \textbf{55.82}  \\ \hline
\end{tabularx}%
\caption{Comparison between Reflexion and Matrix with \texttt{gpt-4o-mini} and \texttt{gpt-4o} as backbone for verbal feedback and optimization. In both cases, Matrix performs better than Reflexion.}
\label{tab:reflexion_mini}
\end{table}

\section{Conclusion}
This paper explores the application of LLM agents for business document information retrieval, focusing on extracting transport references from invoices. To specialize LLM agents for this domain, we introduce Matrix, a paradigm enabling systematic learning through iterative self-refinement and memory updates. Real-world evaluations show Matrix distills actionable insights, outperforms baselines by large margin, and improves latency, cost, and document processing capability. This work reveals the potential of LLM agents for efficient, scalable enterprise document automation.

\bibliography{aaai25}
\onecolumn
\appendix
\section{Dataset Details}
\label{anon}

The dataset we experiment on contains sensitive data that reveals the customers' business details. Therefore, we are unable to release the original dataset. We will release an anonymized subset of the dataset to facilitate future research in this field. The goal of anonymization is to preserve the original data structure while removing sensitive information. In the following section, we describe the anonymization pipeline.

\subsection{Background and data source}

The dataset used in this study originates from real-world invoices processed by Kuehne+Nagel. These invoices represent a diverse range of business transactions, providing a rich source of data for our transport reference and tracking number processing task.

\subsection{Data collection process}

The data collection was facilitated through the Beyond Work platform, a human/AI collaboration platform for solving tedious work. Specifically, the invoices were processed within a workblock, which is a container for all technology (code, infrastructure, permission etc) needed to solve a distinct process. 

The workblock used to collect this data, targets the process of "Parked Invoices" in which invoices that cannot be automatically processed have to be processed by humans. The workblock uses LLMs to process the invoice and introduces a validation task in which human workers were tasked with validating tracking numbers and transport references extracted from the invoices. This process not only involved identification and verification of the correct information but also required workers to provide explanations for any discrepancies or errors they encountered. This human-in-the-loop approach ensured high-quality, validated data. 

\subsection{Data preparation and anonymization}

Following the validation process, the data underwent several stages of preparation to ensure its suitability for research purposes while maintaining strict privacy standards. 
The anonymization of the data included both pseudonymization of the identifiers we were looking to extract for the invoices, as the original human validation of the identifiers was done on the non-anonymized data. This way we ensure to keep the format of the identifier while removing any traceability to the original numbers. For all other sensitive data we completely anonymized it. 
A separate workblock was authored (the process of using natural language to create workblocks) specifically for the anonymization process. This step was crucial to protect sensitive business information and comply with data privacy regulations.

\subsection{Unstructured data pseudonymization}

For unstructured text data, we specifcally used the claude-3.5-Sonnet language model within the anonymization workblock. This helped us in replacing identifying information with pseudonyms while maintaining the contextual and structural integrity of the data.

\subsection{Structured data anonymization}

For more structured data fields, we applied complete anonymization. This process involved replacing text data with random strings, integer values with random integers, and float values with random floats. This approach ensures that no traceable information remains in the structured fields.

\subsection{Resulting dataset}

The resulting dataset retains the complex structure and challenges of real-world invoice data while being fully anonymized. It preserves the intricate nature of business documents, allowing for realistic evaluation of information extraction techniques while ensuring the confidentiality of the original data sources.

\section{Results on Anonymized Data}
In this section, we present the main results of the proposed method on the released anonymized data. The dataset comprises 127 task instances, with each instance consisting of an invoice document paired with its corresponding transport reference. In real-world scenarios, an invoice may not always contain a valid transport reference, and this characteristic has been intentionally preserved in the anonymized dataset to reflect real-world conditions. Of the 127 documents, 50 contain valid transport references, which form the primary focus of our study\footnote{Data available at \textcolor{magenta}{https://tinyurl.com/3pakk9t4}}.

Our work specifically focuses on cases where transport references are present. While we acknowledge the limited size of the anonymized dataset, this constraint arises from the significant manual effort required for anonymization and rigorous inspection. These measures are essential to ensure compliance with data privacy standards and to prevent the leakage of sensitive information. Our future work involves expanding the anonymized dataset to accommodate scenarios with missing or invalid transport references to provide a broader evaluation of our method.

We randomly select 8 task instances for training, 42 for testing performance. Due to the training set is much smaller than that of the main experiment, the agent's task solving trajectory and the reflections can fit into the optimizer's context window. In the training phase, the agent repeatedly attempts the same batch of tasks and the memory gets updated accordingly.

\begin{figure}[htbp]

    \centering
    \includegraphics[width=0.5\linewidth]{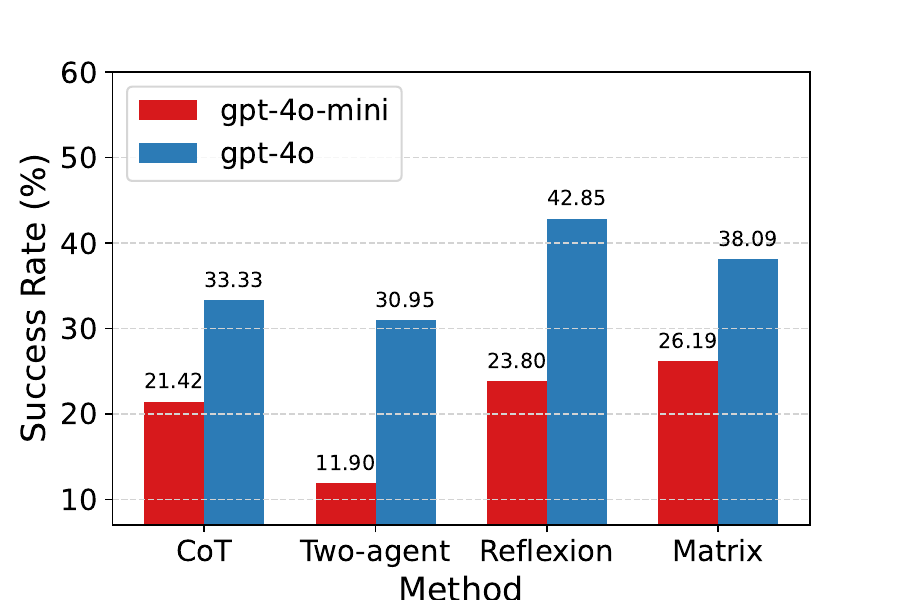}
    \caption{Performance comparison of baselines and Matrix on anonymized dataset. Reflexion performs better than Matrix on \texttt{gpt-4o} agent. In the remaining cases, Matrix performs better.}
\label{appendix:comp}
\end{figure}

We present the performance results on the anonymized dataset in Figure~\ref{appendix:comp}. While Reflexion outperforms Matrix when using \texttt{gpt-4o} as the backbone, Matrix achieves the best performance across all other configurations. However, we observe that Matrix does not exhibit significant improvements over other baseline methods on the anonymized dataset. This is due to the limitations of the training dataset, which is not sufficiently large or diverse to provide a well-generalized representation of the problem space. Specifically, the optimizer is constrained to use the same batch of training tasks across all epochs, resulting in a lack of diversity in the gathered task solving experience. Consequently, the optimizer struggles to capture a comprehensive pattern that accurately reflects the full data distribution.

This highlights a key limitation of Matrix: it requires a substantial amount of training data to effectively model and generalize the patterns within the dataset. Addressing the challenge of agent training under limited data resource constraints remains an avenue for future research.

\section{Implementation Details of baselines}
\label{appendix:implementation}
\subsection{Chain-of-thought}
CoT is often used as the default way of prompting LLM. We follow this guideline and prompt the LLM to think step by step and analyze the dataset before generating the final answer.

\subsection{Two-agent}
The vanilla two agent system is implemented by AG2\footnote{https://github.com/ag2ai/ag2} (formerly AutoGen)~\citep{wu2023autogen}. In the system, an assistant agent backed by LLM is responsible for analyzing the task environment, reason, and make decisions. A user proxy agent receives the content generated by assistant agent, execute the python code provided, and automatically provides the feedback. For a detailed description of the system, please refer to the official documentation\footnote{https://ag2ai.github.io/ag2/docs/tutorial/introduction}.

\subsection{Reflexion}
Reflexion proposes to improve language agents using verbal feedback. For each task, the agent reflects on their performance and store these reflections in memory to make better decisions in the next trial. The system maintains specific memories for each individual task or instance. We change the ReAct agent of the original paper into the two-agent system we investigate, and run reflexion on each task. The maximum number of trials is set to 7 to ensure a fair comparison. The prompt for generating reflection is identical to that of their original Github repository\footnote{https://github.com/noahshinn/reflexion/blob/main/alfworld\_runs/generate\_reflections.py\#L15}.

\section{Limitations and future work}
As shown in Section \ref{weak}, the performance of Matrix strongly depends on the capability of backbone models. Weaker models cannot distill actionable insights from experiences and correctly guide future task solving attempts. 

The process of agent training requires a number of task instances that are representative of the full data distribution. Comparing performance of Matrix on full data and the anonymized data, we discover that the method requires larger training data to reach strong performance. The anonymized data contains little samples that cannot fully reflect the dataset distribution, therefore Matrix tends to underperform. How to identify a subset of the most important and influential samples (i.e. coreset selection~\citep{xia2022moderate,xia2024refined}) for training is an open question for agent training. One potential solution is to leverage AutoML based methods~\citep{wang2021flaml,zheng2023ddpnas,zhang2023targeted,zhang2024hypertime}.

\end{document}